%% file: main.tex
\documentclass[10pt,twocolumn,letterpaper]{article}

\usepackage{cvpr}              %

\input{preamble}
\definecolor{cvprblue}{rgb}{0.21,0.49,0.74}
\usepackage[pagebackref,breaklinks,colorlinks,allcolors=cvprblue]{hyperref}

\usepackage[utf8]{inputenc} %
\usepackage[T1]{fontenc}    %
\usepackage{color}
\usepackage{xcolor}
\definecolor{citecolor}{HTML}{0071BC}
\definecolor{linkcolor}{HTML}{ED1C24}
\definecolor{verylightgray}{gray}{0.9}
\usepackage{amsfonts}       %
\usepackage{nicefrac}       %
\usepackage{microtype}      %
\usepackage{xcolor}         %
\usepackage[breakable]{tcolorbox}
\usepackage{multirow}
\usepackage{amsmath}
\usepackage{bm} %
\usepackage{mathtools} %
\usepackage{graphicx}
\usepackage{wrapfig}
\usepackage{float}
\usepackage{xspace}
\usepackage{multirow}
\usepackage{colortbl}

\usepackage{caption}
\usepackage{subcaption}

\usepackage[utf8]{inputenc} %
\usepackage[T1]{fontenc}    %
\usepackage{url}            %
\usepackage{booktabs}       %
\usepackage{amsfonts}       %
\usepackage{nicefrac}       %
\usepackage{microtype}      

\usepackage{xcolor} %

\usepackage{graphicx}

\usepackage{bbding}
\usepackage{amssymb}
\usepackage{graphicx}
\usepackage{wrapfig}
\usepackage{float}
\usepackage{xspace}
\usepackage{multirow}
\usepackage{colortbl}
\usepackage{makecell}
\usepackage{pifont}
\usepackage{tcolorbox}
\usepackage[utf8]{inputenc} %
\usepackage[T1]{fontenc}    %
\usepackage{url}            %
\usepackage{booktabs}       %
\usepackage{amsfonts}       %
\usepackage{nicefrac}       %
\usepackage{microtype}      
\usepackage{xcolor}         %
\usepackage{mdframed}
\usepackage{graphicx}

\usepackage{bbding}
\usepackage{amssymb}
\usepackage{graphicx}
\usepackage{wrapfig}
\usepackage{float}
\usepackage{xspace}
\usepackage{multirow}
\usepackage{colortbl}
\usepackage{makecell}
\usepackage{caption}
\usepackage{subcaption}
\usepackage{pifont}
\usepackage{tcolorbox}
\usepackage{enumitem}
\makeatletter
\DeclareRobustCommand\onedot{\futurelet\@let@token\@onedot}
\def\@onedot{\ifx\@let@token.\else.\null\fi\xspace}
\definecolor{hpink}{HTML}{F7E1ED}
\definecolor{kellygreen}{rgb}{0.3, 0.73, 0.09}
\definecolor{alizarin}{rgb}{0.82, 0.1, 0.26}

\def\eg{\emph{e.g}\onedot}

\def\modelname{\textbf{EMA}}

\usepackage[capitalize]{cleveref}
\makeatother
\definecolor{green1}{rgb}{0.13, 0.55, 0.13}
\makeatletter
\DeclareRobustCommand\onedot{\futurelet\@let@token\@onedot}
\def\@onedot{\ifx\@let@token.\else.\null\fi\xspace}

\def\eg{\emph{e.g}\onedot}

\makeatother

\def\paperID{8249} %
\def\confName{CVPR}
\def\confYear{2025}

\title{Efficient Motion-Aware Video MLLM}

\author{Zijia Zhao$^{1, 2}$, Yuqi Huo$^{3}$, Tongtian Yue$^{1, 2}$, 
Longteng Guo$^{1, 2}$ , \\ Haoyu Lu$^{4}$ , Bingning Wang$^{3}$, Weipeng Chen$^{3}$, Jing Liu$^{1, 2}$\footnotemark[2] \\ \\
$^1$Institute of Automation, Chinese Academy of Sciences \\
$^2$School of Artificial Intelligence, University of Chinese Academy of Sciences \\
$^3$Baichuan Inc. 
$^4$Renmin University of China 
}
 
\begin{document}
\maketitle
\footnotetext[2]{Corresponding author.}
\input{sec/0_abstract}    
\input{sec/1_intro}
\input{sec/2_related}

\input{sec/3_method}

\input{sec/4_exp}
\section{Acknowledgement}
This research is supported by Artificial Intelligence-National Science and Technology Major Project (2023ZD0121200) and the National Natural Science Foundation of China (6243000159, 62102416), and the Key Research and Development Program of Jiangsu Province under Grant BE2023016-3.

{
    \small
    \bibliographystyle{ieeenat_fullname}
    \bibliography{main}
}

\input{sec/X_suppl}

\end{document}

%% file: sec/0_abstract.tex
\begin{abstract}
Most current video MLLMs rely on uniform frame sampling and image-level encoders, resulting in inefficient data processing and limited motion awareness. To address these challenges, we introduce \modelname, an \textbf{E}fficient \textbf{M}otion-\textbf{A}ware video MLLM that utilizes compressed video structures as inputs. We propose a motion-aware GOP (Group of Pictures) encoder that fuses spatial and motion information within a GOP unit in the compressed video stream, generating compact, informative visual tokens. By integrating fewer but denser RGB frames with more but sparser motion vectors in this native slow-fast input architecture, our approach reduces redundancy and enhances motion representation. Additionally, we introduce MotionBench, a benchmark for evaluating motion understanding across four motion types: linear, curved, rotational, and contact-based. Experimental results show that \modelname\  achieves state-of-the-art performance on both MotionBench and popular video question answering benchmarks, while reducing inference costs. Moreover, \modelname\  demonstrates strong scalability, as evidenced by its competitive performance on long video understanding benchmarks.
\end{abstract}

%% file: sec/1_intro.tex
\section{Introduction}
\label{sec:intro}
Understanding video content is essential for comprehensive world modeling. Recent advancements in multimodal large language models (MLLMs) for video understanding leverage uniform frame sampling~\cite{zhang2023videollama, lin2023videollava, jin2024chatunivi, li2023llama, maaz2023videochatgpt, yang2022frozenbilm, zhang2023llamaadapter, li2023videochat, song2023moviechat} and apply vision encoders~\cite{zhai2023siglip, radford2021learning}, which are originally designed for image processing, to analyze each video frame. This approach concatenates frame features temporally before inputting them into the LLM for complex analysis. Here, the vision encoder captures spatial semantics per frame, while the LLM primarily handles temporal comprehension. 
However, \textit{does this framework genuinely achieve comprehensive video understanding?}

\begin{figure}[t!]
    \centering
    \includegraphics[width=\linewidth]{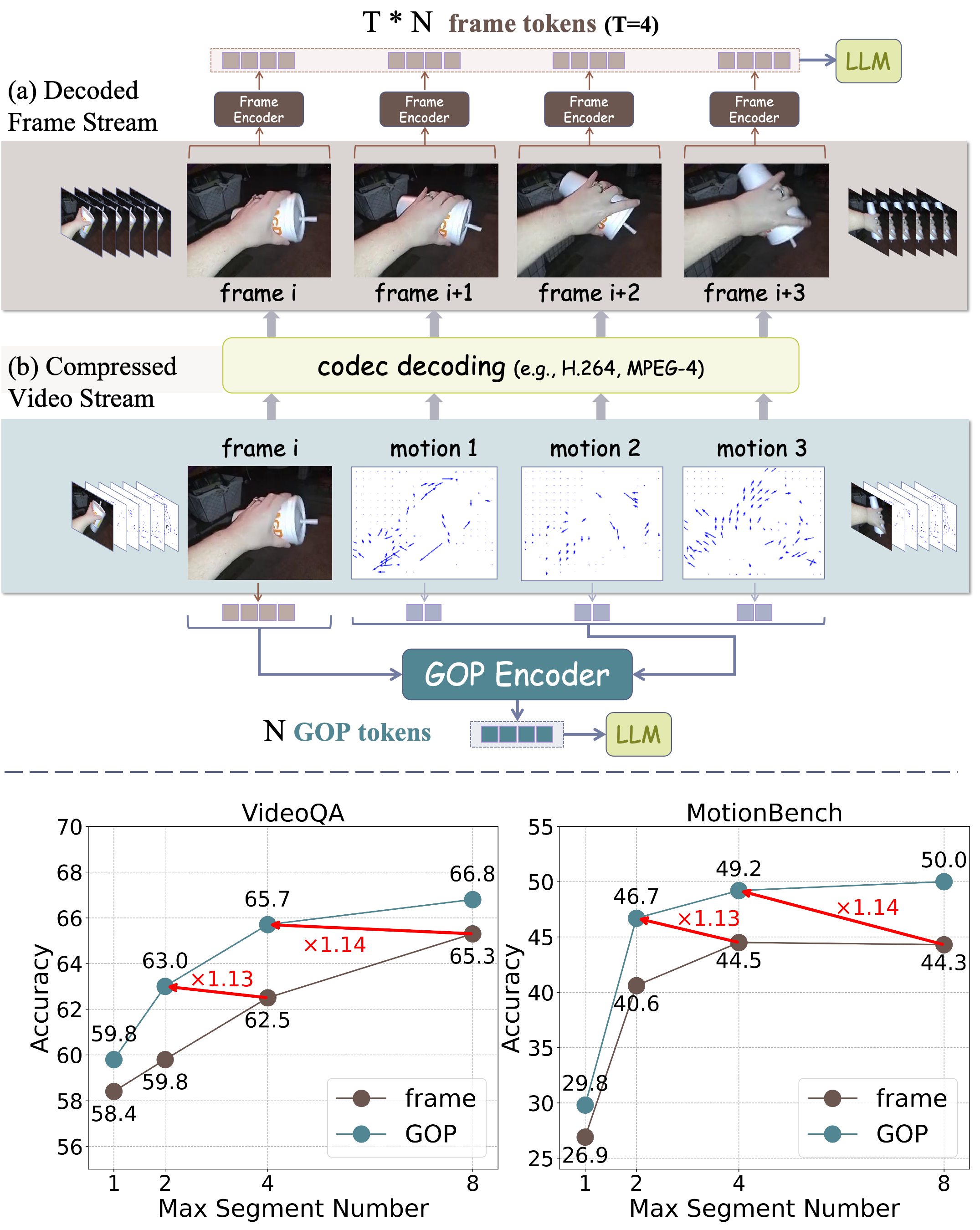} %
    \caption{Comparison of sampling from decoded frames versus the GOPs (Group Of Pictures) from compressed video stream. Compressed video encoding generates tokens at only 1/T the length of sampled frames for the same clip, while capturing motion information more directly. It also shows greater efficiency on Video-QA (average of MSVD-QA and MSRVTT-QA~\cite{xu2017video}) and MotionBench within our \modelname\  framework, achieving higher accuracy with less inference time (\textcolor{red}{red arrow} shows inference speed up).}
    \label{illsfig}
    \vspace{-0.5cm}
\end{figure}

The core difference between images and videos is that, unlike static images, videos contain motion information, which previous methods struggle to handle effectively. As illustrated in \cref{illsfig}, consider a short segment from a video showing a person’s hand holding a cup. When using uniform frame sampling, challenges arise: sampling a single frame (\eg, frame \(i\)) may make it difficult, even for humans, to discern the cup’s movement direction. To better capture motion, denser sampling is needed ((\eg, frame \(i\sim i+T\))), producing a longer sequence of visual tokens (\eg, \(T \times N\)), which increases inference costs.
Some methods \cite{li2023llama, xu2024pllava} improve efficiency in long video analysis by reducing the number of visual tokens per frame, often at the cost of frame-level semantic detail. Other approaches \cite{xu2024slowfastllava, zhang2024llavavideo, du2024exploring} exploit the redundancy between neighboring frames, employing a slow-fast encoding strategy to process large numbers of frames while compressing tokens through dynamic pooling. These methods mainly address the redundancy in individual frame features due to the similarity between adjacent frames in the video sequence. However, such redundancy can also be reduced prior to visual encoding.

We approach video encoding from a different perspective of video stream inputs. In conventional electronic video codecs (\eg, MPEG-4, H.264), videos are stored with high compression ratios, significantly reducing redundancy. The compressed video stream holds enough information to fully reconstruct the video into frames, showing it contains what's needed to understand the video. 
Compression is achieved by segmenting video data into a limited set of RGB frames (I-frames) and a larger set of \textit{sparse} motion vectors and residuals (P/B-frames). Most frames in a video sequence are reconstructed by referencing an I-frame and applying corresponding motion vectors and residuals.
Using compressed videos~\cite{li2020slow, chen2022mm, huang2021self,huo2021compressed, huo2020lightweight,shen2023accurate, jin2024videolavit} as input (with dense RGB values in I-frames and sparse motion vectors in P/B-frames) naturally forms a slow-fast input structure. This architecture minimizes redundancy compared to a fully decoded frame stream, lessening the need to compress redundant frames. Notably, the motion vectors inherently track the trajectories of macroblocks within each frame, making this compression-domain structure highly effective for capturing motion dynamics.

Building on this concept, we introduce an \textbf{E}fficient \textbf{M}otion-\textbf{A}ware video understanding MLLM \modelname, which separately encodes key frames and motion frames from compressed video streams. Inspired by modern codecs' frame decoding, we propose a GOP encoder module to encode and integrate RGB and motion data, which segments video into GOP units and fuses spatial and motion semantics into the same number of visual tokens as a single frame while containing more information inside a video clip. The GOP feature sequence is then fed into a large language model for video understanding via a two-stage training. To evaluate motion understanding, we created MotionBench, a benchmark assessing four common types of motion: linear, curved, rotational, and contact-based.
Our model achieved superior performance on both MotionBench and public benchmarks, including MSVD-QA~\cite{xu2017video}, MSRVTT-QA~\cite{xu2017video}, and ActivityNet-QA~\cite{yu2019activitynetqa}, while demonstrating reduced inference costs. Furthermore, when evaluated on long video understanding benchmarks, such as VideoMME~\cite{fu2024videomme}, the model maintained competitive performance. These experimental results indicate that our approach not only lowers the inference cost of Video MLLMs but also enhances motion capture and overall video understanding.

Our contribution can be concluded as:
\begin{itemize}
    \item We introduce \modelname, an efficient motion-aware video model that leverages compressed video structures as slow-fast input to enhance motion capture and minimize input redundancy.
    \item Recognizing the lack of benchmarks dedicated to motion understanding, we develop MotionBench, a custom evaluation benchmark that assesses video motion understanding.
    \item Our model achieves competitive results on both MotionBench and public datasets with reduced inference costs. Additionally, it performs well on long video understanding benchmarks, demonstrating good context scalability.
\end{itemize}

%% file: sec/2_related.tex
\section{Related Works}
\label{sec:related}
\subsection{Video MLLMs}

Recent developments in multimodal large language models have extended their applications to video data \cite{yang2022frozenbilm, zhang2023videollama, song2023moviechat}, where uniform frame sampling and frame-by-frame encoding are commonly employed. Early models, such as VideoChat \cite{video-chat}, VideoChatGPT \cite{maaz2023videochatgpt}, and Valley \cite{luo2023valley}, rely on mean pooling to aggregate frame-level features before feeding them to LLMs, which limits temporal comprehension. Furthermore, some advanced models have introduced specific enhancements to capture dynamic content; for instance, Chat-UniVi \cite{jin2024chatunivi} utilizes dynamic token clustering, and BT-Adapter \cite{liu2023btadapter} incorporates lightweight adapters to improve image LLMs for video data. VideoChat2 \cite{li2023videochat} further replaces CLIP \cite{radford2021learning} with a dedicated video encoder \cite{li2023unmasked}, achieving better temporal modeling but at the cost of complex, multi-stage training. Video-LaVIT \cite{jin2024videolavit} decomposes video content into keyframes and motion discrete vectors, capturing temporal dynamics through a tokenization strategy.

By contrast, Our approach fundamentally departs from prior methods by directly leveraging compressed video data to efficiently capture both spatial and motion information in a unified architecture. Unlike methods that rely on separate tokenizer decomposition (\eg, Video-LAVIT) or pooling strategies (\eg, VideoChatGPT), our model processes video through a GOP encoder, simultaneously integrating keyframes and motion vectors within each GOP unit. This design reduces redundancy at the data input level, resulting in a more compact streamlined architecture that enhances both efficiency and motion comprehension.
\subsection{Compressed Video Understanding}

The use of motion in video processing originates from early MPEG-4 codec-based research designed to enhance action recognition tasks \cite{li2020slow, chen2022mm, huang2021self}. Foundational methods like \cite{li2020slow} employ a slow-i-fast-p architecture that prioritizes keyframes for spatial context while using sparse motion frames to efficiently capture temporal dynamics. Later, a self-supervised video representation learning approach \cite{huang2021self} introduced context-motion decoupling for improved temporal understanding without explicit labels. Similarly, Mm-ViT \cite{chen2022mm} integrates multi-modal inputs, including motion data, within a video-transformer framework to improve action recognition. Subsequent works utilizing H.264 compression techniques (\eg, \cite{huo2021compressed, huo2020lightweight}) further optimized motion vector tokenization, demonstrating efficient motion data processing across extended video clips. Beyond action recognition, motion-aware frameworks have expanded into other video tasks. In the captioning domain, \cite{shen2023accurate} employs motion vectors for real-time video descriptions, while VideoComposer \cite{wang2024videocomposer} incorporates motion control for compositional video synthesis. Motion information has also been adapted for semantic segmentation \cite{hu2023efficient} and prompt tuning \cite{li2023compressed}.

Our work is a pioneering approach in integrating compressed video-based motion encoding within multimodal large language models for video understanding. By unifying motion and keyframe data in a GOP structure, our method captures spatiotemporal semantics with fewer tokens, achieving more efficient processing and marking a significant advancement over prior frame-sampled approaches.

%% file: sec/3_method.tex
\section{Method}
In this section, we mainly introduce the specific design of the proposed model \modelname. At the same time, considering the current lack of motion capability detection for video MLLMs, we also manually construct a motion benchmark MotionBench.
\subsection{Compressed-domain Video Input}

\begin{figure}[t!]
    \centering
    \includegraphics[width=\linewidth]{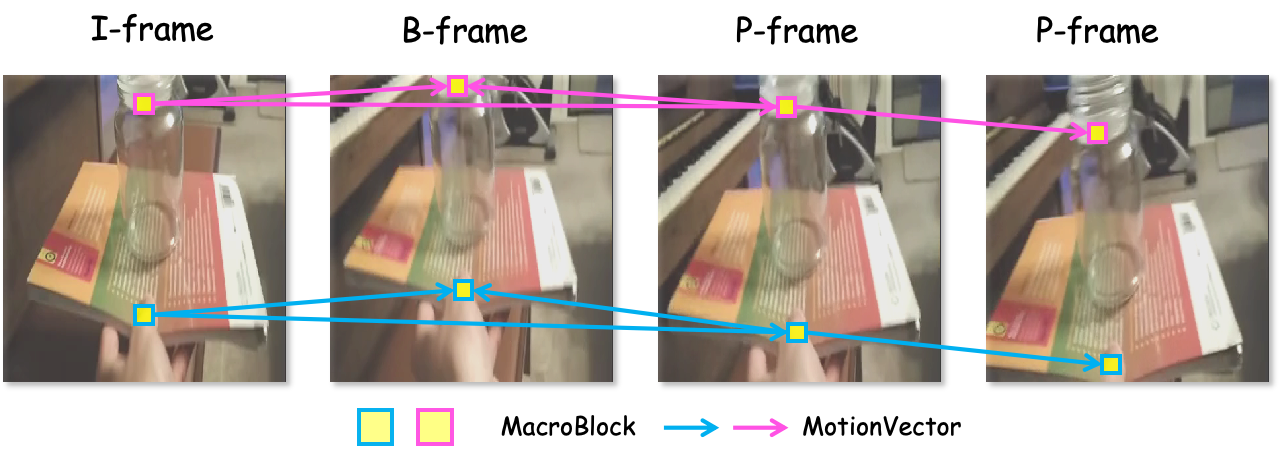} %
    \caption{Description of the H.264 codec modes for a single GOP (Group of Pictures). P/B-frames are decoded sequentially in the decoding order. Decoding relies on motion vectors, which record the movement of each macroblock in the current frame relative to those in reference frames.}
    \label{H.264ills}
    \vspace{-0.5cm}
\end{figure}

We use compressed-domain videos as input to our video understanding model, which exhibit less redundancy compared to frame sampling inputs. In modern codecs (\eg, H.264), video frames are categorized into three types: I-frames (intra-coded frames), P-frames (predictive-coded frames), and B-frames (bipredictive-coded frames). As shown in \cref{H.264ills}, I-frames are encoded independently and rely solely on the current frame, while P and B frames depend on adjacent reference frames , using these references along with the motion vectors and residuals stored at the current position to predict the current frame. 

Motion Vector (MV) represents the displacement of a macroblock from the current frame to a reference frame. If a macroblock at position \( (x, y) \) in the current frame corresponds to a macroblock at position \( (x', y') \) in the reference frame, the motion vector is defined as:
\begin{align}
MV(x, y) = (x' - x, y' - y)
\end{align}

Residuals represent the prediction errors between the actual and predicted macroblocks. If \( P(x, y) \) is the actual pixel value in the current frame, and \( P'(x, y) \) is the predicted value from the reference frame with motion vector, the residual is:
\begin{align}
\Delta(x, y) = P(x, y) - P'(x, y)
\end{align}
Thus, the decoded P/B frame is predicted using the reference frame \( I_{ref} \), the motion vector \( MV \), and the residual \( \Delta \):
\begin{align}
    I_{P/B} = \text{Pred}(I_{ref}, MV) + \Delta
\end{align}
Frames are grouped into Group of Pictures (GOP) units, each containing one I-frame and several P/B-frames, with each GOP decoding independently of others.

In our method, we segment the video input into several GOP units based on the codec structure of the source video. Within each GOP, we select the RGB values from the I-frame and uniformly sample a fixed number of forward motion vectors from the P/B-frames. This forms the compressed video segment input, defined as follows:
\begin{align}
    \centering
    \text{Input}_{\text{Video}}& = \ 
    \underbrace{\left[I_1, MV_{(1,1)}, MV_{(1,2)}, \dots, MV_{(1,M)}\right]}_{\bm{\text{GOP}_1}}, \dots, \notag \\
    &\ 
    \underbrace{\left[I_N, MV_{(N,1)}, MV_{(N,2)}, \dots, MV_{(N,M)}\right]}_{\bm{\text{GOP}_N}}
\end{align}
Here, $N$ represents the GOP segment number, $M$ the motion vector frame number within a GOP, $I_k$ the keyframe in the $k$-th GOP, and $MV_{(k,t)}$ the $t$-th motion vector frame in the $k$-th GOP.
This GOP structure, compared to the decoded frame sequence, does not require additional decoding from the original data stream. Additionally, since motion vectors and residuals are more sparse than the decoded frames, this data format has less redundancy.
\begin{figure*}[t!]
    \centering
    \includegraphics[width=\linewidth]{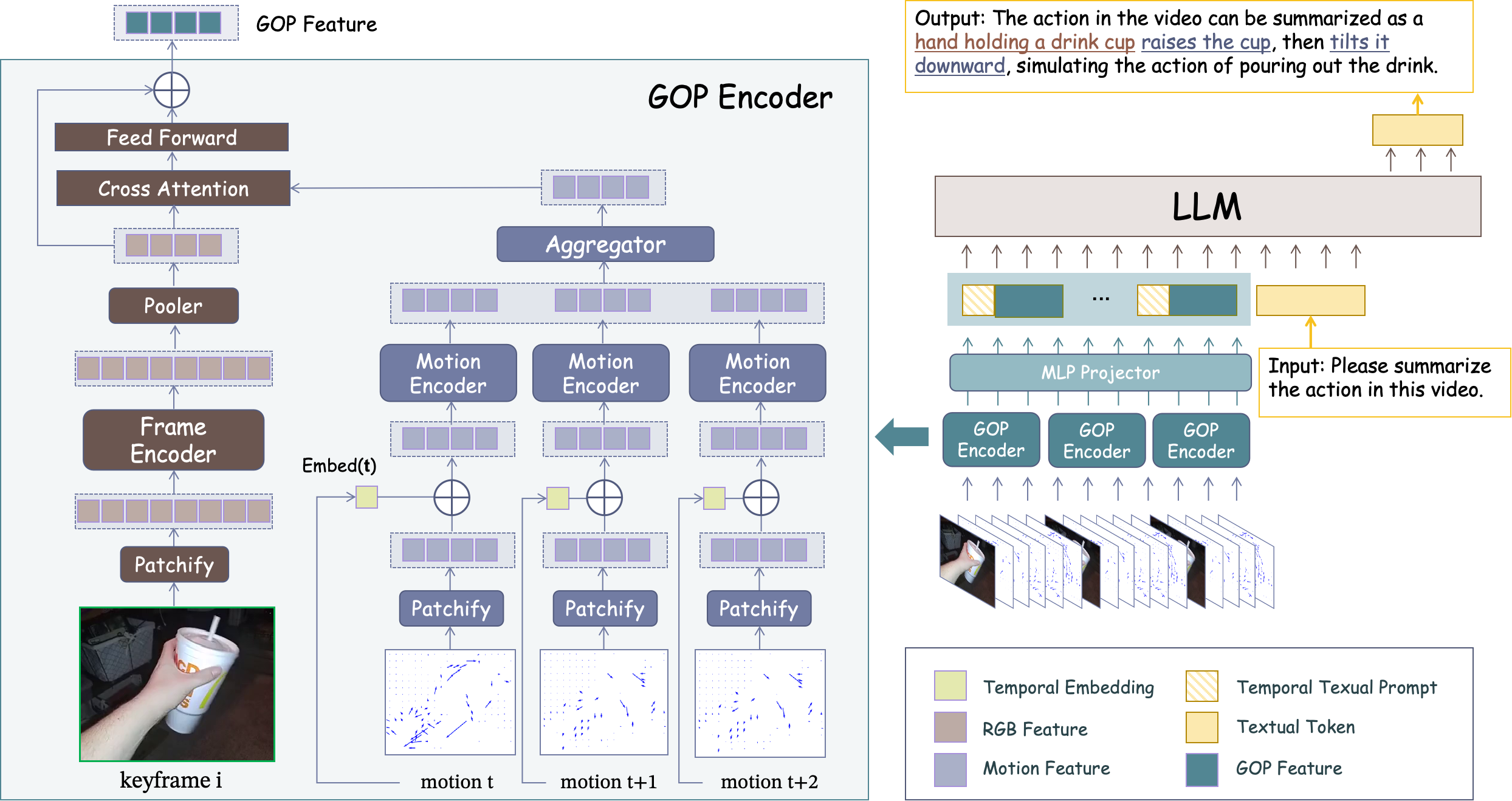} %
    \caption{An illustrative diagram of the overall model architecture. The compressed-domain video stream is divided into GOPs (Groups of Pictures), and each GOP is encoded using our designed GOP encoder. After concatenation, the encoded GOPs are input into the LLM along with text instructions. On the left side of the figure is the detailed structure of the GOP encoder, which decouples frame and motion encoding. It fuses the frame features with the aggregated motion feature sequence to produce a fixed-length GOP feature containing both spatial and motion information.}
    \vspace{-0.5cm}
\end{figure*}

\subsection{Motion-aware GOP Encoder}
We primarily encode the RGB and motion information within the GOP, utilizing the sparser compressed input for more efficient encoding compared to decoded frames. Our goal is to represent a video segment with the same number of tokens as a single frame, while capturing both spatial and motion information. To achieve this, we employ a decoupled encoding strategy, integrate motion position awareness, and apply attention fusion to build a motion-aware GOP encoder.

In each GOP, we decouple key frame and motion vector encoding using two separate encoders. A vision encoder, $\text{Enc}_{I}$, encodes the RGB values of the I-frame $I_k$, with dimensions $(h, w, 3)$. An adaptive pooling layer is then applied to reduce the number of visual tokens.
\begin{equation}
    \bm{F}^I_k = \text{Pooling} \big( \text{Enc}_I(I_k) \big)
\end{equation}
For motion vectors $MV_{(k,t)}$, with a macroblock size of 4, the dimensions are $(h//4, w//4, 2)$. Notably, learnable positional embeddings are added to the patchified motion vectors to preserve temporal information, as the reconstruction in the H.264 codec maintains temporal order.
\begin{align}
    \bm{F}^{MV}_{(k,t)} = &\ \text{Enc}_{MV} \Big( \text{Patchify}(MV_{k,t}) + \notag \\
    &\ \text{PosEmbed}(t) \Big), \quad t = 1, \dots, M
\end{align}

Then, we design a fusion module to fuse the key frame RGB feature $\bm{F}^I_k$ and  motion feature sequence $\bm{F}^{MV}_{(k,t), t=1,...,M}$ within an independent GOP segment $\bm{\text{GOP}_k}$. The resulting vision tokens in the GOP feature encapsulate both spatial semantics and motion information, while maintaining the same token count as a single frame feature.
We aggregate the motion features using mean pooling along the temporal dimension.
\begin{equation}
    \bm{F}^{MV}_k = \text{Aggregator} \Big( \big[ \bm{F}^{MV}_{(k,t)} \big]_{t=1}^{M} \Big)
\end{equation}
Then, we use a fusion layer to append motion feature $\bm{F}^{MV}_k$ on the spatial RGB feature $\bm{F}^I_k$. This layer includes a cross-attention mechanism, a feed-forward layer, and a residual bottleneck. The fusion layer aligns spatial objects with their corresponding motion semantics.
\begin{align}
    \bm{F}^{\text{attn}}_k = &\text{Attention}\big(Q = \bm{F}^I_k, K, V = \bm{F}^{MV}_k\big) \\
    &\bm{F}^{\text{GOP}}_k = \text{FFN}\big(\bm{F}^{\text{attn}}_k\big) + \bm{F}^I_k
\end{align}

\subsection{Overall Architecture}
We divide the input videos into several GOP segments in compressed domain and encode each GOP into a fixed length of GOP features with the GOP encoder. An MLP modality projector then maps these GOP features to visual tokens. These visual tokens are then concatenated in temporal order, with a temporal prompt  $TP$ indicating the time sequence of each GOP. 
\begin{align}
    \bm{X}_V = \big[ TP_1, \text{MLP}(\bm{F}^{\text{GOP}}_1), \dots, TP_N, \text{MLP}(\bm{F}^{\text{GOP}}_N) \big]
\end{align}
where $TP$ is a tokenized textual phrase indicating the temporal information of current video segment $\bm{\text{GOP}_k}$, such as "Segment k".
These visual tokens, along with human-provided textual instructions, are fed into a LLM to perform advanced video understanding tasks.

\subsection{Training Strategy}
\label{trainingstrategy}
We adopt a two-stage training strategy to training our model: visual-text alignment and visual instruction tuning.

 \textbf{Stage 1: Visual-Text Alignment.} In this stage, we aim to build modality alignment between visual context and its textual description.
We use image-text dataset LLaVA-558k~\cite{llava} and video-text dataset Valley-702k~\cite{luo2023valley} as training datasets. For images, we treat them as a static video represented by a single GOP, with a blank motion vector indicating no motion. The image or video is input into the model, and the model is tasked with generating the corresponding caption. The loss function for stage 1 is defined as follows:
\begin{equation}
    \text{Loss}_{\text{stage1}} = -\sum_{\substack{i=1 \\ i \in \bm{X}_{\text{cap}}}}^L \log P_{\theta^{*}}(x_i \mid \bm{X}_V, \bm{X}_{\text{cap}, <i})
\end{equation}
where \( L \) denotes the sequence length, \( \bm{X}_{\text{cap}, <i} \) represents the caption tokens preceding the \(i\)-th token, and \( \theta^{*} \) refers to the trainable parameters in the motion encoder, GOP fusion layer, and modality projector.

 \textbf{Stage 2: Visual Instruction Tuning.}
In this stage, we train the model with visual instructions and responses to better understand human instructions in visual contexts. We collected instruction tuning data from image-text instruction dataset Cauldron~\cite{idefics2} and video instruction dataset VideoChat2-IT~\cite{li2024mvbench}, resulting in a total of 2.4M samples.
The loss function for stage 2 is defined as follows:
\begin{equation}
    \text{Loss}_{\text{stage2}} = -\sum_{\substack{i=1 \\ i \in \bm{X}_{\text{ans}}}}^L \log P_\theta \left( x_i \mid \bm{X}_V, \bm{X}_{\text{ins}, <i}, \bm{X}_{\text{ans}, <i} \right)
\end{equation}
where \( L \) is the sequence length, \( \bm{X}_{\text{ans}, <i} \) and \( \bm{X}_{\text{ins}, <i} \) represent the tokens from the answer and instruction sequences preceding the \(i\)-th token, and \( \theta \) denotes model parameters.

\textbf{Motion Encoder Warmup.} 
Since the motion encoder is a randomly-initialized transformer, we apply a warm-up phase to its parameters to facilitate faster convergence. To achieve this, we use Something-to-Something V2 (SSV2) dataset~\cite{goyal2017something2} and employ a supervised learning approach. Only motion vectors extracted from compressed videos are used as input, without any RGB information. An external classification head is added to the motion encoder, and it is trained with cross-entropy loss against the ground-truth action labels prior to Stage 1.

 \begin{figure}[t!]
    \centering
    \label{motionbench}
    \includegraphics[width=\linewidth]{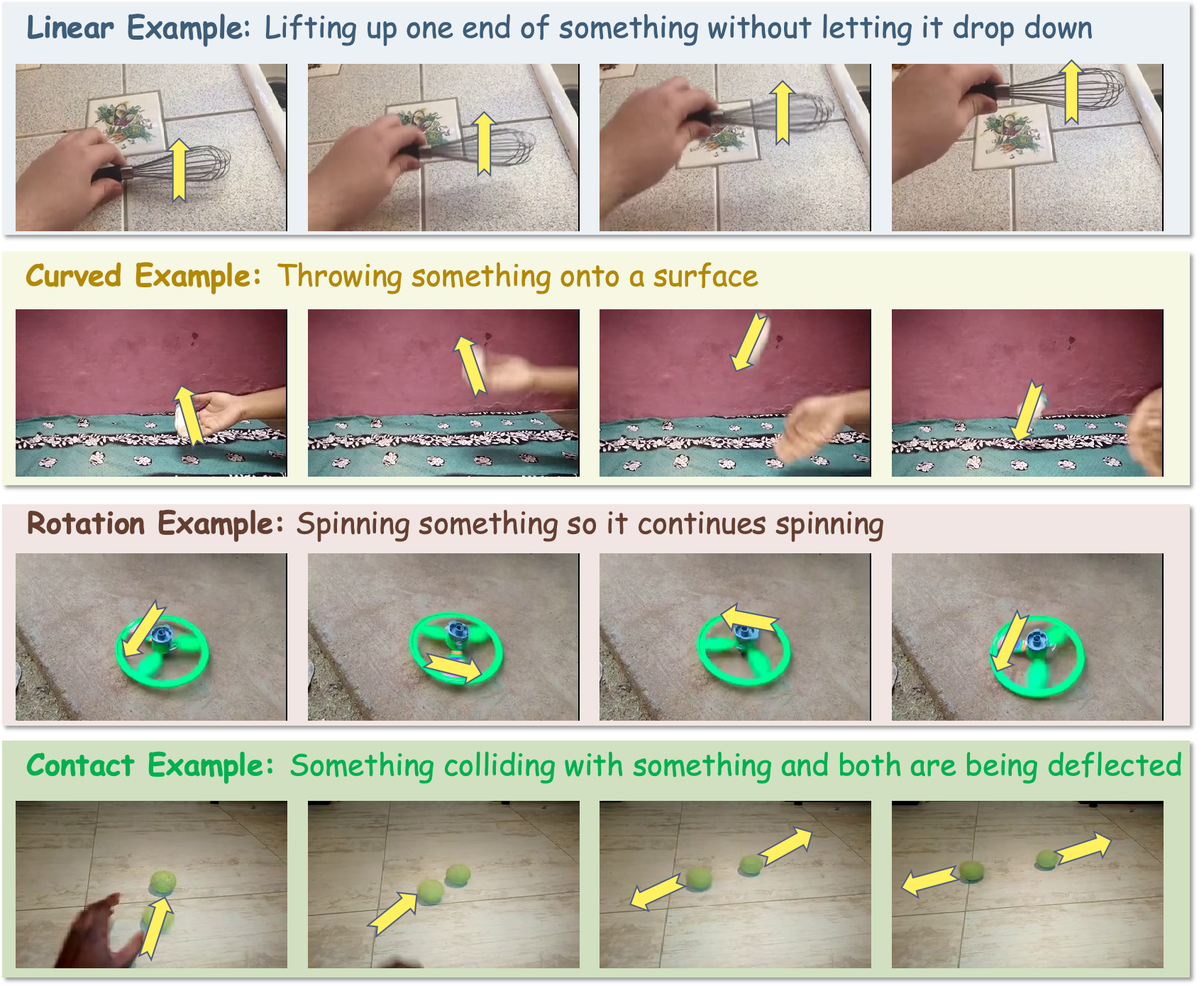} %
    \caption{Examples of data from MotionBench. We show four types of examples: linear, curved, rotation, and contact. \textcolor{yellow}{Yellow arrows} in the video frames indicate the motion trajectories of the same specified object. Different trajectory patterns correspond to different data types.}
    \vspace{-0.7cm}
\end{figure}

 \subsection{MotionBench}
We observed that existing video benchmarks lack specialized metrics for evaluating motion information, which hinders precise assessment of a model’s sensitivity to motion dynamics. To address this gap, we introduce a custom benchmark, MotionBench, designed to evaluate video models' understanding of motion patterns. We collect 2.3k videos from various sources~\cite{goyal2017something2, kay2017kinetics, soomro2012ucf101}. Each video in MotionBench contains a single distinct motion pattern without any scene transitions. Specifically, we collected 114 unique motion patterns, categorizing each video under one of these patterns manually as shown in \cref{motionbench}. Based on the characteristics of motion trajectories, we classified the test data into four main categories: Linear (Lin.), Curved (Cur.), Rotation (Rot.), and Contact (Con.). The characteristics of each category are detailed below:
 \begin{itemize}
     \item \textbf{Linear}: Motion occurring in a consistent direction, such as a toy car moving from one side to another.
     \item \textbf{Curved}: Motion following a curved trajectory, like a ball thrown from a height toward the ground.
     \item \textbf{Rotation}: Clockwise rotational motion, such as a spinning CD or plate.
     \item \textbf{Contact}: Motion involving two objects making contact, often with compression, such as the collision of two balls.
 \end{itemize}

During testing, we employed a multiple-choice format for each video, presenting one correct option alongside three carefully crafted incorrect options. These incorrect options were intentionally selected as the most confusable motion patterns relative to the ground truth among the 114 classes, ensuring a rigorous evaluation of the model’s ability to accurately comprehend motion dynamics. We report the multi-choice QA accuracy as the performance on MotionBench.

%% file: sec/4_exp.tex
\begin{table*}[t!]
\centering
\caption{Comparison with SOTA models on public videoQA benchmarks: MSVD-QA \cite{chen2011collecting}; MSRVTT-QA \cite{xu2016msr};  ActivityNet-QA \cite{caba2015activitynet}.  Maximum values are in \textbf{bold}.}
\vspace{-0.2cm}
\resizebox{0.65\linewidth}{!}{  
\begin{tabular}{lccccccccccc}
  \toprule
  \multirow{2}{*}{\bf Method} & \multicolumn{2}{c}{\bf MSVD-QA} & & \multicolumn{2}{c}{\bf MSRVTT-QA} & & \multicolumn{2}{c}{\bf ActivityNet-QA} \\ \cline{2-3} \cline{5-6} \cline{8-9}
  & Acc & Score & & Acc & Score & & Acc & Score \\
  \midrule
  FrozenBiLM~\cite{yang2022frozenbilm}  & 32.2 & -- & & 16.8 & -- & & 24.7 & -- \\
  VideoLLaMA ~\cite{zhang2023videollama} & 51.6 & 2.5 & & 29.6 & 1.8 & & 12.4 & 1.1 \\
  LLaMA-Adapter ~\cite{zhang2023llamaadapter}  & 54.9 & 3.1 & & 43.8 & 2.7 & & 34.2 & 2.7 \\
  VideoChat ~\cite{li2023videochat} & 56.3 & 2.8 & & 45.0 & 2.5 & & 26.5 & 2.2 \\
  Video-ChatGPT ~\cite{maaz2023videochatgpt} & 64.9 & 3.3 & & 49.3 & 2.8 & & 35.2 & 2.7 \\
  BT-Adapter ~\cite{liu2023btadapter} & 67.5 & 3.7 & & 57.0 & 3.2 & & 45.7 & 3.2 \\
  LLaMA-VID ~\cite{li2023llama} & 69.7 & 3.7 & & 57.7 & 3.2 & & 47.4 & 3.3 
  \\

 Chat-UniVi~\cite{jin2024chatunivi}  & 65.0 & 3.6 & & 54.6 & 3.1 & & 45.8 & 3.2 
  \\
 Video-LLaVA~\cite{lin2023videollava}  & 70.7 & 3.9 & & 59.2 & \textbf{3.5} & & 45.3 & 3.3 
  \\
  MovieChat~\cite{song2023moviechat}  & 75.2 & 3.8 & & 52.7 & 2.6 & & 45.7 & 3.4 
  \\
  Video-LaVIT~\cite{jin2024videolavit}  & 73.2 & 3.9 & & \textbf{59.3} & 3.3 & & 50.1 & 3.3 
  \\

  \midrule

\modelname &  \textbf{75.8} & \textbf{4.1} &  & 58.5 & \textbf{3.5} & & \textbf{52.1} & \textbf{3.5}    \\
  \bottomrule
\end{tabular} 
}
\label{tab:video_perf} 
\end{table*}
\begin{table*}[t!]

\centering
\caption{Performance comparison with state-of-the-art models on long video tasks VideoMME~\cite{fu2024videomme}. Maximum values are in \textbf{bold}.}
\vspace{-0.2cm}
\resizebox{0.8\linewidth}{!}{
\begin{tabular}{lccccccccc}
\toprule
\multirow{2}{*}{\bf{Method}} & \multicolumn{4}{c}{\bf{VideoMME  w/o subs}} & & \multicolumn{4}{c}{\bf{VideoMME w subs}} \\
\cline{2-5} \cline{7-10}
                       & short   & medium   & long   & \bf{Overall} & & short  & medium  & long  & \bf{Overall}  \\
                       \midrule
Video-LLaVA ~\cite{lin2023videollava}           & 45.3    & 38.0     & 36.2   & 39.9 &     & 46.1   & 40.7    & 38.1  & 41.6     \\
Qwen-VL-Chat ~\cite{bai2023qwenvl}           & 46.9    & 38.7     & 37.8   & 41.1 &     & 47.3   & 40.4    & 37.9  & 41.9     \\
ST-LLM  ~\cite{liu2025stllm}               & 45.7    & 36.8     & 31.3   & 37.9  &    & 48.4   & 41.4    & 36.9  & 42.3     \\
VideoChat2-Mistral  ~\cite{li2024mvbench}               & 48.3   & 37.0     & 33.2   & 39.5  &    & 52.8  & 39.4  & 39.2& 43.8     \\
Chat-UniVi-v1.5 ~\cite{jin2024chatunivi}       & 45.7    & 40.3     & 35.8   & 40.6  &    & 51.2   & 44.6    & 41.8  & 45.9     \\
LongVA   ~\cite{zhang2024longva}              & 61.1    & 50.4     & 46.2   & 52.6   &   & 61.6   & 53.6    & 47.6  & 54.3     \\
\midrule
\modelname                    &  65.8   &   50.2    &  44.3  & \bf{53.4}  & & 65.3  &    55.4    &  54.3         & \bf{58.4}    \\
\bottomrule
\end{tabular}}
\vspace{-0.2cm}
\end{table*}
\section{Experiments}
\subsection{Implementation Details}
We employ Qwen2-7B~\cite{yang2024qwen2} as the LLM backbone. The frame encoder is derived from SigLIP-so400m~\cite{zhai2023siglip}. The motion encoder consists of a two-layer transformer with two channels. Details are provided in the Appendix.

We re-encode the videos using the H.264 codec with a maximum keyframe interval of 8 and a frame rate of 4 fps. During the training process, we uniformly sample groups of pictures (GOPs) from each video, limiting the maximum number of GOPs to 8. Each GOP consists of one RGB I-frame and a uniformly sampled set of motion vectors, with a maximum of 8 motion vectors. For image data, we treat each image as a static video segment; thus, an image GOP includes the RGB image and a zero motion vector. All RGB frames in a GOP are resized to 384 × 384 pixels, while all motion vectors are resized to 96 × 96 pixels, corresponding to a macroblock size of 4 in H.264 encoding. To enhance training and inference efficiency, the features of the RGB frames are processed using adaptive pooling with a 3×3 kernel. Each batch contains a random combination of images and videos, with data samples packed in a sequence limited to a maximum of 4096 tokens.

In the first stage, we train for 1000 steps with a batch size of 128, where only the motion encoder, GOP fusion layer, and modality projector are trainable. In the second stage, we train for 2400 steps with the same batch size, but all model parameters are fine-tuned. Our experiments are conducted on 16 NVIDIA A100 GPUs, and the training hyperparameters are detailed in the Appendix.

\begin{table*}[t!]
\centering

\caption{Efficiency analysis under different model, video segment type and number. We report the inference time and maximum visual token count for MLLM generation. We use model-\( f \) and model-\( g \) to denote models trained with frame and GOP inputs, respectively. Model-D refers to the final model used as \modelname. Maximum values are in \textbf{bold}, and second-highest values are \underline{underlined}.}
\vspace{-0.2cm}
\renewcommand{\arraystretch}{1.1}
\resizebox{\linewidth}{!}{  
\begin{tabular}{c|cc|c|c|ccccc|cc}
\toprule
\multirow{2}{*}{Models} & Segment & Max Seg.& MSVD       & MSRVTT     & \multicolumn{5}{c|}{MotionBench} & \#Visual& Inference \\
                          & Type    & Number  & Acc./Score & Acc./Score & Lin.   & Cur.   & Rot.  & Con. & Avg. & Tokens   & Time   \\
                          \midrule
Video-LLaVA~\cite{lin2023videollava}& frame   & 8       &             70.7 / 3.9&             \textbf{59.3} / 3.5&        36.1&       31.2&       32.0&       44.5&          36.0&      2048&   391.4ms \\
LLaMA-VID~\cite{li2023llama}& frame   & 8       &             69.7 / 3.7&             57.7 / 3.2&        34.1&       35.8&       37.0&       43.0&          37.5&    2048  & 273.7ms\\
\midrule
model-\( f \)1& frame   & 1       &            70.1 / 3.7&            46.7 / 3.1&        21.9&        33.9&       25.7&       23.1&          26.9&      81&   93.5ms\\
\rowcolor{verylightgray} model-\( g \)1& GOP     & 1       &            71.2 / 3.7& 48.4 / 3.2   &        26.3&        35.5&       30.0&       27.4&          29.8&      81&  94.0ms\\
model-\( f \)2& frame   & 2       &            71.0 / 3.7&            48.6 / 3.2&        42.6&        39.1&       44.3&       34.5&          40.6&      162&   97.0ms\\
\rowcolor{verylightgray}model-\( g \)2& GOP     & 2       &            72.4 / 3.8&  53.6 / 3.4  &        52.6&        41.8&       47.7&       45.5&          46.7&      162&   97.4ms\\
model-\( f \)4& frame   & 4       &            72.4 / 3.8&            52.6 / 3.3&        53.5&        39.8&       45.7&       36.3&          44.5&      324&   110.8ms\\
\rowcolor{verylightgray}model-\( g \)4& GOP     & 4       &            \underline{74.3} / 4.0& 57.0 / 3.5 &        59.8&        42.7&       47.7&       46.6&          \underline{49.2}&      324& 111.5ms\\
model-\( f \)8& frame   & 8       &            74.0 / 4.0&            56.7 / 3.4&        53.5&        40.4&       43.3&       37.1&          44.3&      648&    127.1ms\\
\rowcolor{verylightgray}model-\( g \)8 (\modelname)& GOP     & 8       &            \textbf{75.8} / 4.1&  \underline{58.5} / 3.5&        60.1&        43.6&       49.3&       47.0&          \textbf{50.0}&    648&  127.1ms\\
\bottomrule
\end{tabular}}
\label{efftable}
\vspace{-0.3cm}
\end{table*}
\subsection{Quantitative Evaluation}
\subsubsection{Performance on Video-QA Benchmarks}
We conduct a quantitative evaluation on public video question-answering benchmarks, including MSVD-QA~\cite{xu2017video}, MSRVTT-QA~\cite{xu2017video}, and ActivityNet-QA~\cite{yu2019activitynetqa}. For this evaluation, we follow a standard approach~\cite{maaz2023videochatgpt}, utilizing GPT-3.5-turbo to evaluate answers, providing both accuracy and score metrics. 
\modelname\  outperforms various models using uniform frame sampling methods, such as Video-LLaVA~\cite{lin2023videollava}, LLaMA-VID~\cite{li2023llama}, and Chat-UniVi~\cite{jin2024chatunivi}. Video-LaVIT~\cite{jin2024videolavit} also leverages an MPEG-4 codec to decouple keyframes and motion vectors. However, it requires a higher number of input tokens compared to \modelname, as it simply concatenates keyframe and motion tokens without integrating them through fusion. \modelname\ achieves superior performance over Video-LaVIT on MSVD-QA and ActivityNet-QA and achieves comparable results on MSRVTT-QA.

\subsubsection{Extension on Long Video Benchmarks}
Our approach can also be adapted for long video understanding tasks. We increased the model's maximum GOP count to 32 and trained it with the same setting. We evaluate on long video benchmark VideoMME~\cite{fu2024videomme}. Our results demonstrate that the GOP method is effective for longer video tasks, achieving competitive performance with most video models.

\subsection{Efficiency Analysis}
To evaluate the efficiency of our method, we report task performance, maximum input token count, and average inference time per sample in \cref{efftable}. The inference time refers to the average MLLM generation time per sample.

\modelname\  achieves superior performance on MSVD-QA and MotionBench compared to previous methods such as Video-LLaVA~\cite{lin2023videollava} and LLaMA-VID~\cite{li2023llama}. Additionally, our model uses fewer visual tokens (648 vs. 2048) and demonstrates lower inference time (×3.08 compared to Video-LLaVA, ×2.14 compared to LLaMA-VID), highlighting its efficiency in video understanding.

We also compare model performance and inference time across different video segment types in \cref{efftable}. Specifically, we compare two input types: video frames, commonly used in previous models, and GOPs, used by \modelname. All models are trained under identical settings, except for differences in segment type and count.
Using GOPs as the video segment unit results in only a negligible increase in MLLM inference time compared to direct frame input, while maintaining the same visual token count. However, model-\(g\) achieve a significant performance boost over model-\(f\) using frame input with the same segment count, with a comparable inference time overhead. Notably, on MotionBench, GOPs outperform frames due to their superior ability to capture motion information.
Additionally, we observe that GOP encoding outperforms models with denser frame sampling, as shown in \cref{illsfig} and \cref{efftable}, while also requiring less inference time (×1.13 comparing model-\(g\)2/\(f\)4 and ×1.14 comparing model-\(g\)4/\(f\)8). This demonstrates that our approach not only improves inference efficiency but also reduces redundancy in video inputs for video MLLMs.
\subsection{Ablation Study}
In this section, we conduct comprehensive ablation studies on module configuration within the GOP encoder. All models are trained on the same dataset and with identical hyperparameter settings, using a maximum of 4 GOPs.

\begin{table*}[t!]
\centering
\caption{Ablation experiments on the method design. We compared the construction of the fusion module, the encoding modes of motion positional information, and the model training strategies. We evaluated the performance of all models on MSVD-QA~\cite{xu2017video}, MSRVTT-QA~\cite{xu2017video}, and MotionBench. Exp 0 is the final setting we utilized. Maximum values are in \textbf{bold}, and second-highest values are \underline{underlined}.}
\vspace{-0.1cm}
\resizebox{0.83\linewidth}{!}{  
\begin{tabular}{c|ccc|c|c|ccccc}
\toprule
\multirow{2}{*}{Exp.} & Fusion  & Motion  & Motion & MSVD-QA       & MSRVTT-QA     & \multicolumn{5}{c}{MotionBench} \\
                          &    Type   &   Position   & Warmup& Acc./Score & Acc./Score & Lin.   & Cur.   & Rot.  & Con. & Avg.  \\
\midrule
\rowcolor{verylightgray} 0                         & crossattn                    & pre-fusion                         &  \ding{51}                            & \underline{74.3} / 4.0& \underline{57.0} / 3.5&  59.8&  42.7&  47.7&  46.6&  \textbf{49.2}\\
\midrule
1                         & w/o motion                          & -                        &  \ding{55}                            &            72.4 / 3.8&  52.6 / 3.3&        53.5&        39.8&       45.7&     36.3&   44.5\\
2                         & add                          & pre-fusion                        &  \ding{51}                            &            73.2 / 3.9&  56.6 / 3.3&        55.2&        41.4&       44.3&     43.1&   46.2\\
3                         & concat                       & pre-fusion                       &  \ding{51}                            &            73.4 / 3.9&  56.7 / 3.3&        56.1&        42.3&       45.6&     44.1&   47.0\\
4& w/o agg.                       & pre-fusion                       &  \ding{51}                            &            74.1 / 4.0&  \textbf{57.2} / 3.5&        58.8&        42.5&       48.6&     46.2&   \underline{49.0}\\
\midrule
5& crossattn                    & post-fusion                      &  \ding{51}                            &   \textbf{75.0} / 4.1&  56.9 / 3.4&        57.8&        42.3&       46.3&     45.0&   47.9\\
6& crossattn                    & no pos                          &   \ding{51}                           &  73.2 / 4.0&  56.4 / 3.3&        54.1&        40.3&       43.2&     39.6&   44.8\\
\midrule
7& crossattn                    & pre-fusion                         &  \ding{55}                         &  73.2 / 4.0&  56.3 / 3.2&        54.5&        40.3&       44.6&    40.8&   45.6\\
\bottomrule
\end{tabular}}
\label{abla}
\vspace{-0.3cm}
\end{table*}
\paragraph{Fusion module architecture in GOP encoder}
We evaluate different fusion strategies in the GOP encoder in Exp (0-4) of \cref{abla}. The add method simply sums the motion features and frame features within a GOP. The concat method concatenates motion and frame features along the last dimension. The crossattn method utilizes a cross-attention module to integrate motion information into the frame embeddings. Additionally, we experiment with using the entire set of motion features as the key/value inputs in the cross-attention module without first aggregating them through a pooling layer.
The results indicate that the crossattn method yields a slight improvement over the add and concat methods on Video-QA tasks. Notably, it provides greater gains on MotionBench, suggesting that our fusion module architecture effectively integrates motion information with the semantics of static frames. We also observe that a pooling aggregator achieves performance comparable to that of the full motion sequence while reducing computational costs. Consequently, we adopt a cross-attention fusion module with a pooling aggregator as our final design.

\paragraph{Motion Encoder Architecture.} 
Since motion vectors are sparse relative to RGB frames, we employ a lightweight transformer to encode them. In \cref{motionlayer}, we present model performance across different numbers of layers in the motion encoder. The results show a slight performance improvement with more than two layers, suggesting that the sparse motion vectors do not require a complex module for effective encoding. Consequently, we select a two-layer transformer as the motion encoder.

\begin{table}[t!]
\centering
\caption{Ablation study on motion encoder layer number. 'w/o enc' means encoding without motion encoder. Maximum values are in \textbf{bold}, and second-highest values are \underline{underlined}.}
\vspace{-0.2cm}
\label{motionlayer}
\renewcommand{\arraystretch}{1.1}
\resizebox{0.83\linewidth}{!}{ 
\begin{tabular}{cccc}
\toprule
Encoder & MSVD-QA    & MSRVTT-QA  & MotionBench \\
Layer   & Acc./Score & Acc./Score & Avg. Acc.   \\
\midrule
w/o enc       &            72.4 / 3.8&            52.6 / 3.3&             44.5\\
1       &            73.4 / 3.9&            56.4 / 3.4&             47.2\\
\rowcolor{verylightgray} 2       &            \underline{74.3} / 4.0&            \textbf{57.2} / 3.5&             \textbf{49.2}\\
3       &            \textbf{74.4} / 4.0&            \textbf{57.2} / 3.5&             48.8\\
4       &            74.1 / 4.0&            \underline{57.0} / 3.5&            \underline{49.0}\\
\bottomrule
\end{tabular}}
\vspace{-0.3cm}
\end{table}

\paragraph{Motion Position Encoding}
Motion vectors represent the movement from the previous frame to the current frame, creating a natural chronological order for the motion vector sequence within a GOP. To encode the temporal index of each motion vector, we use a learnable position embedding within the GOP. In Exp (0, 5-6) of \cref{abla}, we compare various methods for incorporating temporal position embeddings into motion features. We examine three approaches: No Pos (Exp 6) encodes motion features without temporal position information,
Pre-Fusion (Exp 0) adds learnable position embeddings to the motion encoder inputs, and Post-Fusion (Exp 5) adds learnable position embeddings to the motion encoder outputs.
Our results show that the Pre-Fusion method achieves the best performance, indicating that early fusion supports more effective temporal modeling of motion sequences.

\paragraph{Motion Warmup Strategy}
Since the motion encoder is randomly initialized while the frame encoder is initialized with the pretrained vision backbone SigLIP-so400m~\cite{zhai2023siglip}, we employ a warm-up strategy to enhance the motion encoder. Specifically, we use motion vectors from the Something2Something-v2 dataset~\cite{goyal2017something2} along with their corresponding labels, applying a supervised training method with cross-entropy loss. This warm-up approach significantly improves model performance, as evidenced by the comparison between Exp 0 and 7 in \cref{abla}.

\section{Conclusion}
In this work, we introduced \modelname, a efficient motion-aware video understanding model for comprehensive video analysis by leveraging the native slow-fast structure of compressed video data. Our approach employs a motion-aware GOP encoder to effectively fuse spatial and motion information within compressed video units. This innovative encoding method integrates dense RGB spatial semantics with sparse motion dynamics, thereby reducing redundancy and inference costs while preserving robust video understanding capabilities.
Additionally, we present MotionBench to address the gap in motion-focused evaluation of video models. Our model demonstrated superior performance across both public video benchmarks and MotionBench with less inference cost, underscoring its effectiveness in video comprehension. We believe that \modelname\ and MotionBench represent important steps toward more efficient and accurate video understanding, with the potential to inspire future research that further investigates compressed video structures and motion-centric analysis within multimodal language models. We hope our research will inspire future advancements in the efficiency and motion awareness of video MLLMs.

%% file: sec/X_suppl.tex
\appendix
\onecolumn
\section{Model Details}
\label{modeldetails}
Our model consists of three main components: the GOP encoder, the MLP projector, and a large language model (LLM). In the GOP encoder, we leverage a pre-trained SigLIP-so400m~\cite{zhai2023siglip} as the RGB frame encoder. Simultaneously, we utilize a custom-designed transformer as the motion encoder to extract motion information.

Given that the motion vectors extracted from compressed video streams are represented as a discrete list \( L \):
\begin{align}
    L &= \text{List}\left[(x_{\text{src}}^i, y_{\text{src}}^i, x_{\text{dst}}^i, y_{\text{dst}}^i)\right], \quad i = 1, \ldots, n.
\end{align}
we first transform it into a 2D spatial format \( P \) as follows:
\begin{align}
    &P[x_{\text{src}}^i, y_{\text{src}}^i] = \left(x_{\text{dst}}^i - x_{\text{src}}^i, y_{\text{dst}}^i - y_{\text{src}}^i\right).
\end{align}
Since the motion vector represents displacements for macroblocks of size \(4 \times 4\), and the original frame dimensions are \(h \times w \times 3\), we derive a motion matrix of shape \((h/2, w/2, 2)\). Subsequently, the motion matrix is resized to a fixed resolution of \(96 \times 96\), which is corresponding to the frame resolution \(384 \times 384\). 

We employ a two-layer transformer as the motion encoder, with a hidden size of 256 and 2 channels. The input motion matrix is processed using patches of size \(7 \times 7\). After encoding, the motion feature has the same dimensionality as the frame feature, ensuring consistency across modalities.

For the extracted motion features, we apply temporal pooling along the time dimension to summarize the temporal dynamics. Additionally, to reduce the number of input tokens, we perform adaptive pooling on both the frame features and the motion features. This operation leverages the \texttt{torch.nn.AdaptiveAvgPool2d} module to efficiently compress spatial dimensions while preserving important information.

Subsequently, we employ a fusion layer to integrate the frame and motion information. This fusion process is implemented using a cross-attention layer followed by a feed-forward layer to facilitate modality interaction. Additionally, we incorporate a residual module to retain the input information, ensuring that critical details from both modalities are preserved during the fusion.

\section{Training Hyperparameters}
\label{hyperpara}
During the first-stage training, we freeze the frame encoder and the large language model (LLM) while training the motion encoder, the GOP fusion layer, and the modality projector. A global batch size of 128 is used, and the model is trained for 1000 steps. The motion encoder and projector are optimized with a learning rate of \(1 \times 10^{-4}\), while the remaining components are trained with a learning rate of \(2 \times 10^{-5}\).

In the second-stage training, we unfreeze all model parameters for joint optimization. The learning rates for different components are set as follows: the frame encoder uses a learning rate of \(2 \times 10^{-6}\), the motion encoder uses \(1 \times 10^{-5}\), the projector uses \(1 \times 10^{-4}\), and the remaining components use \(2 \times 10^{-5}\). The training is conducted for 2400 steps with a global batch size of 128.

We employ DeepSpeed ZeRO-2 for distributed training to efficiently handle large-scale models and data. During training, different samples are packed into a single sequence with a maximum length of 4096 for joint optimization, significantly improving training efficiency. The training was conducted on 16 NVIDIA A100 GPUs, with a total training time of approximately 16 hours.

Additionally, the extra motion warmup was conducted on 8 NVIDIA A100 GPUs. During this phase, we utilized a batch size of 1024 for supervised training with a learning rate of \(1 \times 10^{-3}\). The training was performed on the motion vectors of SSV2~\cite{goyal2017something2} training videos for a total of 30 epochs.

\section{MotionBench Details}
\label{motionbenchdetails}
We used the label set from SSV2~\cite{goyal2017something2} as the initial pool of options. Subsequently, we employed GPT-4o as a teacher model to filter these 174 options, extracting 114 classes that could be mapped to our predefined four categories: \textbf{Linear}, \textbf{Curved}, \textbf{Rotation}, and \textbf{Contact}.

To facilitate evaluation, we designed MotionBench as a multi-choice QA task. To increase the task complexity, we identified three hard negative labels for each category, which were included as confusing options in the QA design. GPT-4o assisted in selecting these hard negatives. For example, the following represents a set of confusing options:

\begin{tcolorbox}[breakable,title=Confusing Label Sets]
\label{fig:subtitle}
\renewcommand\baselinestretch{1.5}\selectfont
\small
"Pouring something out of something" \\"Pouring something into something" \\"Pouring something onto something"\\"Pouring something into something until it overflows"
\end{tcolorbox}

In addition, considering the limitations of video types in SSV2, we introduced \cite{soomro2012ucf101, kay2017kinetics} as a supplementary data source. Since these videos lack initial labels, we utilized GPT-4o to generate dense captions for the videos. These dense captions were then matched to candidate categories, with the matching process also conducted by GPT-4o.

After the labeling process was completed, we performed a manual screening of the test videos. During this step, we filtered out incorrectly labeled examples and those that were overly simplistic, such as cases where the answer could be inferred directly from static images.

Finally, MotionBench comprises 4 distinct classes: Linear, Curved, Rotation, and Contact, containing 800, 500, 300, and 700 samples, respectively. We present the accuracy for each individual class as well as the average accuracy across all four classes.
\section{External Ablations}
\label{externalabla}
\subsection{Use of Temporal Prompt}
When feeding GOPs into the LLM, we added a textual temporal prompt for each GOP, which included its time coordinates. We found that this simple approach led to a significant performance improvement on long benchmarks, such as VideoMME~\cite{fu2024videomme}. However, it had a smaller impact on shorter VideoQA benchmarks, such as MSVD-QA~\cite{xu2017video} and MSRVTT-QA~\cite{xu2017video}.
\begin{table*}[h!]
\centering
\caption{Impact of temporal prompt. A simple textual temporal prompt proves beneficial for long video tasks, such as VideoMME, but has a smaller impact on shorter VideoQA tasks, such as MSVD-QA and MSRVTT-QA.}
\begin{tabular}{l|cccc}
\toprule
\multirow{2}{*}{\bf{Model}}                        & \bf{MSVD-QA}      & \bf{MSRVTT-QA}    & \bf{MotionBench} & \bf{VideoMME}        \\
                                              & Acc. / Score & Acc. / Score & Avg.        & w/o sub / w sub \\
                                              \midrule
\modelname                     &              75.8 / 4.1&              58.5 / 3.5&             50.0&                 53.4 / 58.4\\
\modelname~ w/o Temporal Prompt &              75.6 / 4.1&              58.5 / 3.5&             49.8&                 51.2 / 56.7\\
\bottomrule           
\end{tabular}
\end{table*}
\subsection{GOP Token Number}
In \modelname, we employ a 3×3 pooling kernel to reduce the length of GOP tokens by a factor of 9. In this section, we evaluate the impact of this compression strategy across several VideoQA benchmarks. We experiment with different pooling kernel sizes while keeping the rest of the training setup consistent. Our results show that the 3×3 pooling kernel achieves performance comparable to both the 2×2 pooling and no pooling configurations, while benefiting from a significant reduction in token length (1/9 of the original), thereby accelerating inference.

\begin{table*}[h!]
\centering
\caption{Influence of pooling strategy.}
\begin{tabular}{c|c|cccc}
\toprule
\multirow{2}{*}{\bf{Pooling Strategy}}  &   \bf{GOP Token}                  & \bf{MSVD-QA}      & \bf{MSRVTT-QA}    & \bf{MotionBench} & \bf{VideoMME}        \\ &
                         \bf{Number}                     & Acc. / Score & Acc. / Score & Avg.        & w/o sub / w sub \\
                                              \midrule
w/o pooling          &      729   &              76.0 / 4.1&              58.9 / 3.5&             50.2&                 53.6 / 58.9\\
2×2 pooling &    196    &       75.8 / 4.1&              58.4 / 3.5&             49.7&                 53.3 / 58.4\\
\rowcolor{verylightgray}3×3 pooling &   81    &        75.8 / 4.1&              58.5 / 3.5&             50.0&                 53.4 / 58.4\\
4×4 pooling &   49     &       73.6 / 3.9&              56.8 / 3.3&             49.0&                 52.0 / 56.2\\
\bottomrule           
\end{tabular}
\end{table*}
\section{Evaluation Results on More Long Video Benchmark}
\label{longbench}
We evaluate our model's performance on additional long-video benchmarks MLVU~\cite{zhou2024mlvu}, LongVideoBench~\cite{wu2024longvideobench}, and VNBench~\cite{zhao2024needle}. We compare \modelname~ with existing video understanding models. Our model demonstrated outstanding performance across these benchmarks as well.
\begin{table}[h!]
\centering
\caption{Evaluation result on long video benchmarks, MLVU~\cite{zhou2024mlvu}, LongVideoBench~\cite{wu2024longvideobench} and VNBench~\cite{zhao2024needle}}
\begin{tabular}{l|ccc}
\toprule
\multirow{2}{*}{Model}    & MLVU & LongVideoBench & VNBench \\
                          & Dev  & Val            & Overall \\
                          \midrule
                          VideoChat~\cite{li2023videochat}& 29.2& -& -\\
                          VideoChatGPT~\cite{maaz2023videochatgpt}& 31.3& -& 4.1\\
Video-LLaVA~\cite{lin2023videollava}               & 47.3 & 39.1           & 12.4    \\
Video-LLaMA2~\cite{zhang2023videollama}              & 35.5 & -              & 4.5     \\
LLaMA-VID~\cite{li2023llama}                 & 33.2 & -              & 7.1     \\
LLaVA-NeXT-Video~\cite{zhang2024llavanext-video}          & -    & 43.5           & 20.1    \\
ST-LLM~\cite{liu2025stllm}                    & -    & -              & 22.7    \\
LongVA~\cite{zhang2024longva}                    & 56.3 & -              & -       \\
Qwen2-VL-7B~\cite{yang2024qwen2}                    & 55.6 & -              & 33.9      \\
\midrule
\modelname & \bf{57.2} & \bf{47.0}           & \bf{32.6}    \\
\bottomrule
\end{tabular}
\end{table}